\documentclass[sigconf, nonacm]{acmart}

\usepackage{multirow} 

\usepackage{amsmath,amsfonts,bm}

\def\eqref#1{equation~\ref{#1}}

\def\1{\bm{1}}

\def\mI{{\bm{I}}}

\def\mK{{\bm{K}}}

\def\mP{{\bm{P}}}

\def\mR{{\bm{R}}}

\def\mZ{{\bm{Z}}}

\DeclareMathAlphabet{\mathsfit}{\encodingdefault}{\sfdefault}{m}{sl}
\SetMathAlphabet{\mathsfit}{bold}{\encodingdefault}{\sfdefault}{bx}{n}

\AtBeginDocument{%
  }

\begin{document}

\title{TextSplat: Text-Guided Semantic Fusion for Generalizable Gaussian Splatting}

\author{Zhicong Wu}
\authornote{Equal Contribution.}
\affiliation{%
  \institution{Institute of Artificial Intelligence, Xiamen University,}
  \city{Xiamen}
  \country{China}
}

\author{Hongbin Xu}
\authornotemark[1]
\affiliation{%
  \institution{ByteDance Seed,}
  \city{Beijing}
  \country{China}
}

\author{Gang Xu}
\affiliation{%
  \institution{Guangdong Laboratory of Artificial Intelligence and Digital Economy (SZ),}
  \city{Shenzhen}
  \country{China}
}

\author{Ping Nie}
\affiliation{%
  \institution{Peking University,}
  \city{Beijing}
  \country{China}
}
\author{Zhixin Yan}
\affiliation{%
  \institution{School of Future Technology, South China University of Technology}
  \city{Guangzhou}
  \country{China}
}

\author{Jinkai Zheng}
\affiliation{%
  \institution{Hangzhou Dianzi University,}
  \institution{Central Laboratory of Lishui Hospital of Wenzhou Medical University, The First Affiliated Hospital of Lishui University, Lishui People's Hospital,}
  \city{Hangzhou}
  \country{China}
}

\author{Liangqiong Qu}
\affiliation{%
  \institution{University of Hong Kong,}
  \city{Hong Kong}
  \country{China}
}

\author{Ming Li}
\authornote{Corresponding Author.}
\affiliation{%
  \institution{Guangdong Laboratory of Artificial Intelligence and Digital Economy (SZ),}
  \city{Shenzhen}
  \country{China}
}

\author{Liqiang Nie}
\affiliation{%
  \institution{Harbin Institute of Technology (Shenzhen),}
  \city{Shenzhen}
  \country{China}
}

\renewcommand{\shortauthors}{Zhicong Wu et al.}

\begin{abstract}
Recent advancements in Generalizable Gaussian Splatting have enabled robust 3D reconstruction from sparse input views by utilizing feed-forward Gaussian Splatting models, achieving superior cross-scene generalization.
However, while many methods focus on geometric consistency, they often neglect the potential of text-driven guidance to enhance semantic understanding, which is crucial for accurately reconstructing fine-grained details in complex scenes.
To address this limitation, we propose TextSplat---the first text-driven Generalizable Gaussian Splatting framework.
By employing a text-guided fusion of diverse semantic cues, our framework learns robust cross-modal feature representations that improve the alignment of geometric and semantic information, producing high-fidelity 3D reconstructions.
Specifically, our framework employs three parallel modules to obtain complementary representations: the Diffusion Prior Depth Estimator for accurate depth information, the Semantic Aware Segmentation Network for detailed semantic information, and the Multi-View Interaction Network for refined cross-view features.
Then, in the Text-Guided Semantic Fusion Module, these representations are integrated via the text-guided and attention-based feature aggregation mechanism, resulting in enhanced 3D Gaussian parameters enriched with detailed semantic cues.
Experimental results on various benchmark datasets demonstrate improved performance compared to existing methods across multiple evaluation metrics, validating the effectiveness of our framework.
The code will be publicly available.
\end{abstract}

\maketitle

\begin{figure}[t]
\centering
\includegraphics[width=0.93\linewidth]{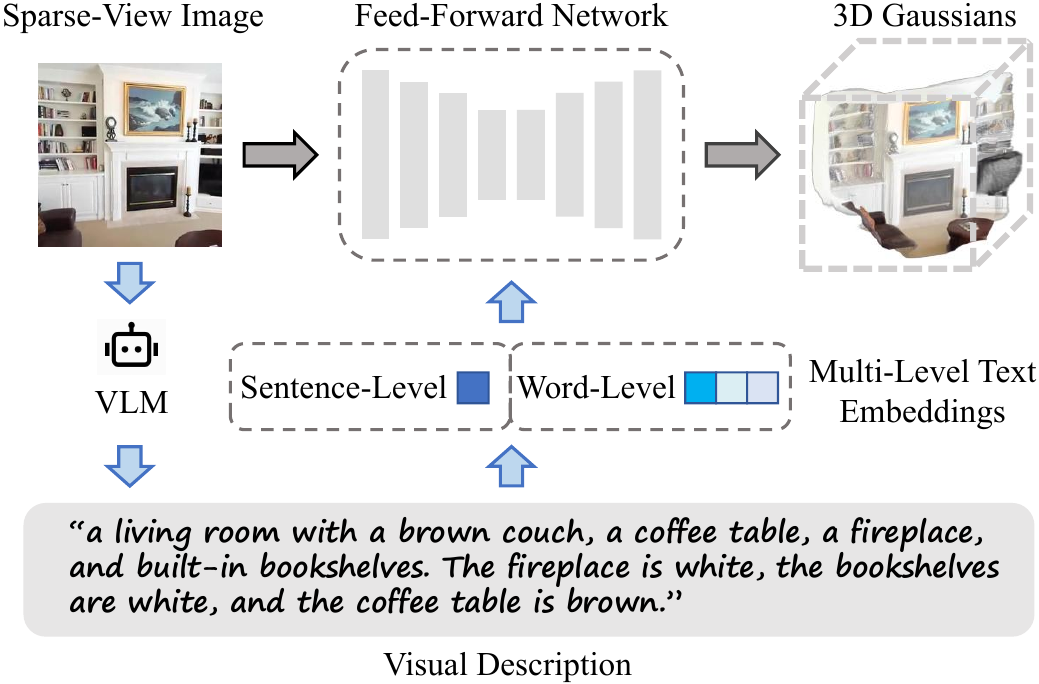} 
\caption{
Overview of the proposed framework, TextSplat, which leverages multi-level text embeddings to facilitate Generalizable Gaussian Splatting from sparse views, as demonstrated on a single-image example.}
\label{first}
\end{figure}

\section{Introduction}

3D Gaussian Splatting (3DGS) \cite{kerbl20233d} has proven to be a highly effective and versatile approach for representing 3D scenes that enables rapid rendering and high-quality reconstruction via rasterization-based techniques.
By bypassing the computationally expensive volumetric sampling process inherent in Neural Radiance Fields (NeRF) \cite{mildenhall2021nerf}, 3DGS significantly reduces computational overhead while preserving visual fidelity.
Based on this foundation, Generalizable Gaussian Splatting (GGS) has been introduced as a novel paradigm for 3D scene reconstruction.
In contrast to traditional scene-specific optimization approaches such as 3DGS and NeRF, GGS leverages pre-trained feed-forward neural networks to instantly predict Gaussian parameters---including position, opacity, covariance, and color—directly from sparse multi-view inputs.
This approach enables robust cross-scene generalization that overcomes the limitations of the conventional per-scene optimization paradigm, making it more practical for real-world applications.

Compared to 3D reconstruction with dense viewpoints, sparse-view 3D reconstruction primarily faces two major challenges: 1) insufficient view coverage leads to uncertainty in geometric structure and the loss of local details, particularly in regions with sparse textures or occlusions; 2) limited overlapping areas between views hinder effective cross-view feature matching and depth estimation, thereby compromising reconstruction accuracy.
Recently, several GGS models~\cite{charatan2024pixelsplat,chen2024mvsplat,zhang2024transplat,tang2024hisplat} have been proposed to tackle these challenges.
Specifically, PixelSplat \cite{charatan2024pixelsplat} breaks new ground by enabling single-pass 3D Gaussian splatting prediction from just two sparse perspective images, surpassing previous NeRF-based methods and setting a new benchmark in GGS.
By constructing the plane-sweeping-based cost volume, MVSplat \cite{chen2024mvsplat} not only achieves accurate positioning of Gaussian centers but also significantly improves the quality of scene predictions.
To address the inherent challenges in multi-view feature matching that critically underpin G-3DGS methodologies, TranSplat incorporates the predicted depth confidence map as a guidance mechanism to enhance local feature matching robustness. 
Despite their success, existing methods neglect the text-driven semantic guidance, leading to unsatisfactory modeling of novel views, especially viewpoints with substantial perspective differences.

Recent advances in text‑driven vision–language models, such as CLIP~\cite{radford2021learning} and Stable Diffusion~\cite{rombach2022high}, have shown that integrating textual cues can markedly improve semantic coherence across tasks like image generation, object detection, and cross‑modal understanding.
Motivated by these findings, we introduce TextSplat, the first text‑driven GGS framework that dynamically fuses multi‑source semantic representations via multi-level text embeddings to enhance geometry–semantic consistency in sparse‑view reconstruction as illustrated in Fig. \ref{first}.
Specifically, our method first employs three parallel modules---Diffusion Prior Depth Estimator, the Semantic Aware Segmentation Network, and the Multi-View Interaction Network---to obtain accurate depth information, detailed semantic information, and refined cross-view features, respectively.
Then, under sentence‑level embedding guidance, the Text-Guided Semantic Fusion Module dynamically integrates these representations via semantic aggregation to enrich multi‑source features and then applies content refinement to recover fine‑grained details lost during fusion.
By integrating high-level semantic descriptions with geometric reasoning, our TextSplat generates robust cross-modal feature representations and achieves precise geometry–semantic alignment.
This is particularly beneficial in sparse-view scenarios, where explicit semantic cues are essential for resolving ambiguities in complex scene structures.
Experimental results on the RealEstate10K~\cite{zhou2018stereo} and ACID~\cite{liu2021infinite} benchmark datasets demonstrate that TextSplat achieves state‑of‑the‑art performance, significantly improving reconstruction quality over existing 3DGS‑ and NeRF‑based methods.

In summary, our contributions are as follows:
\begin{itemize}
\item We introduce TextSplat, the first text‑guided Generalizable Gaussian Splatting framework, integrating textual semantic guidance with multimodal cues to enhance geometry–semantic consistency in sparse‑view reconstruction.
\item We propose the Text-Guided Semantic Fusion Module to achieve multi-source semantic feature integration under the guidance of sentence-level embedding.
\item On public benchmark datasets, TextSplat sets a new state of the art, delivering significantly superior reconstruction quality over existing 3DGS‑ and NeRF‑based methods.
\end{itemize}

\begin{figure*}[t]
\centering
\includegraphics[width=0.86\textwidth]{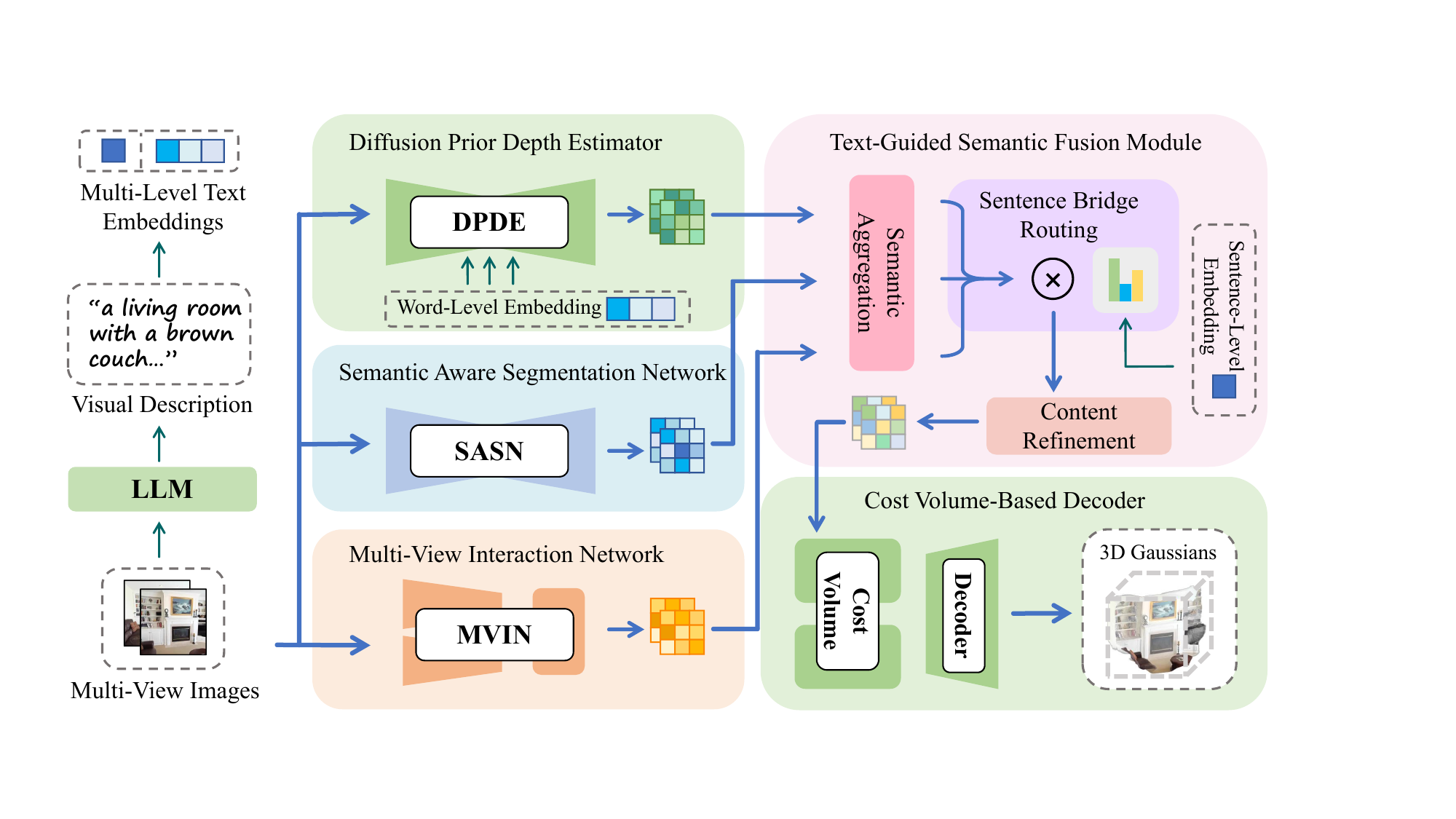} 
\caption{Overview of TextSplat. The framework integrates multimodal cues (diffusion priors, semantic segmentation, cross-view interactions) through a text-guided fusion module to enhance geometry-semantic consistency. Multi-view images and LLM-generated visual descriptions are processed by three parallel networks (DPDE/SASN/MVIN), whose features are dynamically fused via TSFM. The refined features drive cost volume-based decoding for high-fidelity 3D Gaussian parameter prediction. (Example with $K=2$ viewpoints is shown here.)}
\label{network}
\end{figure*}

\section{RELATED WORK}
\subsection{Sparse-View Scene Reconstruction and Synthesis.}

Deep learning has been widely applied in various scenarios such as intelligent transportation \cite{li2021self,li2021exploiting,Yan_2025_CVPR}, cross-modal learning \cite{Liu_2025_CVPR, zhao2025favchat, li2025uni}, privacy protection \cite{li2023dr,li2023stprivacy}, and artificial intelligence generated content \cite{li2024instant3d, liu2024realera}.
Within the domain of computer vision, Multi-View Stereo (MVS) \cite{furukawa2015multi,wang2021multi} stands as a foundational methodology for recovering three-dimensional scene geometries through the analysis of multiple two-dimensional image perspectives.
NeRF (Neural Radiance Fields) \cite{mildenhall2021nerf} is a neural rendering technique that marked a significant breakthrough in computer graphics by representing 3D scenes as continuous volumetric radiance fields, parameterized with neural networks to predict radiance and density at arbitrary spatial locations and viewing directions.
However, this method is computationally intensive \cite{muller2022instant,barron2022mip}, often requiring extensive training times and substantial resources for rendering, especially for high-resolution outputs.
In contrast, 3D Gaussian Splatting (3DGS) \cite{kerbl20233d} emerges as a paradigm-shifting approach that redefines the boundaries of scene representation and rendering. It introduces an explicit scene representation using millions of learnable 3D Gaussians distributed in space, which preserves the strong fitting capability of continuous volumetric radiance fields for high-quality image synthesis. This innovative representation effectively circumvents the computational overhead inherent in NeRF-based methods, such as computationally expensive ray-marching operations and unnecessary calculations in empty spatial regions.

But real-world adoption of original NeRF and 3DGS pipelines faces critical bottlenecks due to their reliance on exhaustive multi-view capture.
This challenge has spurred extensive research interest in sparse-view scene reconstruction and synthesis \cite{charatan2024pixelsplat,chen2021mvsnerf,chen2023explicit,gao2024cat3d,niemeyer2022regnerf,szymanowicz2024flash3d,truong2023sparf,wu2024reconfusion,xu2024murf,zhang2024gs,lin2024phys4dgen}, which seeks to generate high-fidelity 3D representations through limited viewpoint sampling.
Current methodologies addressing sparse-view scenarios are systematically classified into two principal computational frameworks: scene-specific optimization paradigms and neural network-based feed-forward inference architectures.
While the former necessitates time-consuming iterative optimization and exhibits limited generalization due to scene-specific parameter dependencies, the latter demonstrates superior computational efficiency and cross-scenario adaptability through its single-pass neural network pipelines and data-driven prior integration mechanisms.

\subsection{Generalizable Gaussian Splatting}
Generalizable Gaussian Splatting (GGS) emerges as a promising paradigm for sparse-view reconstruction by inheriting 3DGS's explicit 3D Gaussian representation while introducing neural network-driven parameter prediction, achieving fast inference and robust cross-scenario generalization without scene-specific optimization.
Building on this framework, PixelSplat \cite{charatan2024pixelsplat} introduces an epipolar transformer to learn cross-view correspondences and predict depth distributions, achieving single-pass 3D Gaussian splatting from sparse perspective pairs.
To achieve precise 3D Gaussian center localization and enhance geometric accuracy, MVSplat \cite{chen2024mvsplat} proposes a novel cost volume representation constructed through plane sweeping. This representation robustly establishes cross-view correspondences facilitating depth map estimation and Gaussian parameter optimization.
Building upon this foundation, MVSplat360 \cite{{chen2024mvsplat360}} integrates geometry-aware 3D Gaussian Splatting with video diffusion models for high-quality 360° novel view synthesis from sparse inputs.
For sparse-view refinement, TranSplat \cite{zhang2024transplat} adopts a transformer architecture combined with depth-aware deformable matching and monocular depth priors.
DepthSplat \cite{xu2024depthsplat} establishes a unified framework that integrates Gaussian splatting and depth estimation, revealing their inherent task complementarity through bidirectional optimization.
Despite the continuous development of GGS methods, prevailing methodologies predominantly rely on visual information alone for feature extraction and fusion, neglecting the critical role of text-driven guidance in semantic understanding.
Therefore, we first excavated the potential of text-driven semantic fusion approaches in the GGS domain, providing an innovative solution for high-fidelity reconstruction of sparse-view scenes.

\section{Methodology}

\subsection{Problem Definition}

As illustrated in Fig. \ref{network}, we introduce TextSplat, a text-driven Generalizable Gaussian Splatting framework.
Given $K$ sparse-view images $\mathcal{I} = \{\mI_i\}_{i=1}^K$ ($\mI_i \in \mathbb{R}^{H\times W\times 3}$) and their corresponding camera projection matrices $\mathcal{P} = \{\mP_i\}_{i=1}^K$, our TextSplat predicts the comprehensive 3D Gaussian parameters $\mathcal{G}$.
The camera projection matrix on the $i$-th view is: $\mP_i=\mK_i [\mR_i|\bm{t}_i]$, where $\mK_i$, $\mR_i$, and $\bm{t}_i$ are respectively the intrinsic, rotation and translation matrices.
The 3D Gaussian parameters $\mathcal{G}$ include position $\mu$, opacity $\alpha$, covariance $\Sigma$, and color $c$.
These parameters $\mathcal{G}$ support high-fidelity novel view synthesis by leveraging rasterization-based rendering of 3D primitives.

\subsection{Diffusion Prior Depth Estimator}
\label{sec:method:dpde}

To reduce the semantic ambiguity in geometric reasoning, we build the Diffusion Prior Depth Estimator (DPDE) that leverages text-conditional generative priors to bridge 2D observations and 3D scene understanding. 
Similar to \cite{zhao2023unleashing,kondapaneni2024text}, The core architecture of DPDE comprises a visual encoder, a text encoder, a denoising UNet, and a task-specific decoder, as shown in Fig. \ref{three} (a).  


First, the input images $\mathcal{I}$ are fed to the visual encoder to extract the image latent $\mathcal{Z} = \{\mZ_i\}_{i=1}^K$, and the text prompts $\mathcal{T} = \{T_i\}_{i=1}^K$ are fed to the text encoder to obtain the word-level text embedding $\mathcal{C} = \{C_i\}_{i=1}^K$.   
Then, the image latent $\mathcal{Z}$ and text embedding $\mathcal{C}$ are passed through the last step of the denoising process $\epsilon_\theta(\mathcal{Z}, 0, \mathcal{C})$.
Finally, the cross-attention maps and the denoised feature maps are concatenated along the channel dimension and processed by the task-specific decoder to obtain the final diffusion-based depth representations $\mathcal{DF} = \{DF_i\}_{i=1}^K$.

\subsection{Semantic Aware Segmentation Network}
\label{sec:method:sasn}

\begin{figure}[t]
\centering
\includegraphics[width=0.88\linewidth]{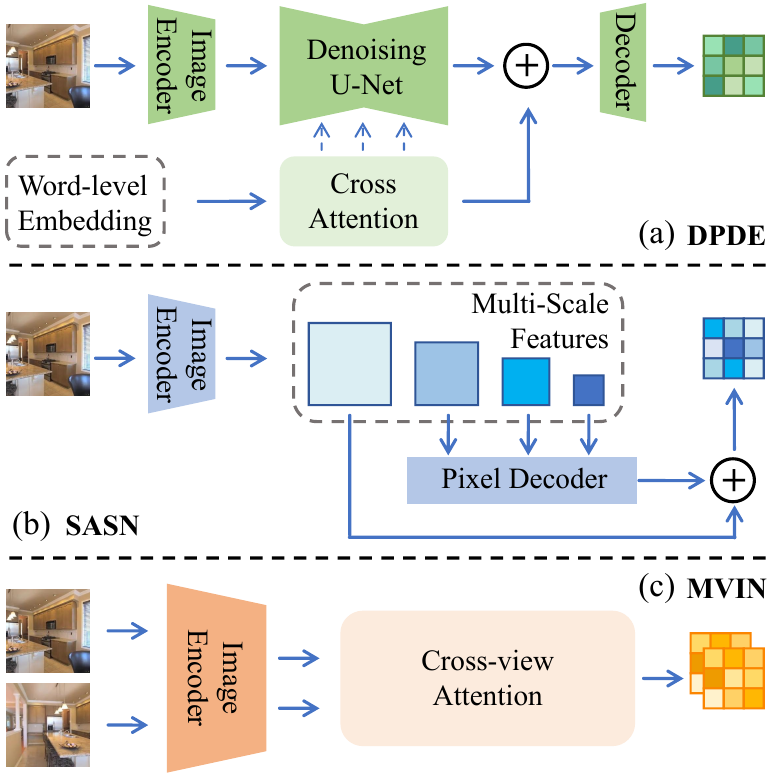} 
\caption{Detailed Architecture of DPDE, SASN, and MVIN. }
\label{three}
\end{figure}

The lack of viewpoints might hinder the holistic scene understanding and fine-grained perception.
To handle this issue, we aim to incorporate the segmentation priors that can provide object-level semantic cues and holistic scene layout information.
Following previous segmentation paradigms \cite{ronneberger2015u,cheng2022masked,zhou2024decoupling,li2024omg}, we design the Semantic Aware Segmentation Network (SASN) with an encoder-decoder architecture, which effectively captures multi-scale semantic cues while preserving spatial-semantic consistency.

Specifically, as shown in Fig. \ref{three} (b), input sparse-view images $\mathcal{I}$ are processed by the Visual Encoder to generate multi-level encoded features (four stages from shallow to deep). The features from the latter three stages are progressively decoded via the pixel decoder to recover spatial details, while the mask feature head fuses the decoded features with the first-stage encoded features to produce final semantic representations $\mathcal{SF} = \{SF_i\}_{i=1}^K$. This design enhances part-level semantic understanding by aggregating both local texture patterns and global contextual information.

\subsection{Multi-View Interaction Network}
\label{sec:method:mvin}

To fully exploit the complementary information across viewpoints and enhance the cross-view consistency for robust 3D reconstruction, we adopt a hybrid CNN-Transformer architecture \cite{xu2023unifying,xu2022gmflow} that models cross-view visual features through explicit inter-view interaction mechanisms.
Specifically, we first employ a shallow ResNet-like CNN to extract per-view image features $\mathcal{CF} = \{CF_i\}_{i=1}^K$, followed by cross-view attention with self- and cross-attention layers to exchange information across different views as shown in Fig. \ref{three} (c).  
Notably, the Transformer architecture specifically employs Swin Transformer's local window attention \cite{liu2021swin} to balance computational efficiency and global context modeling.
This hybrid processing network ultimately generates cross-view semantic features $\mathcal{MF} = \{MF_i\}_{i=1}^K$ that integrate both local texture patterns and global geometric consistency.

\begin{figure}[t]
\centering
\includegraphics[width=0.38\textwidth]{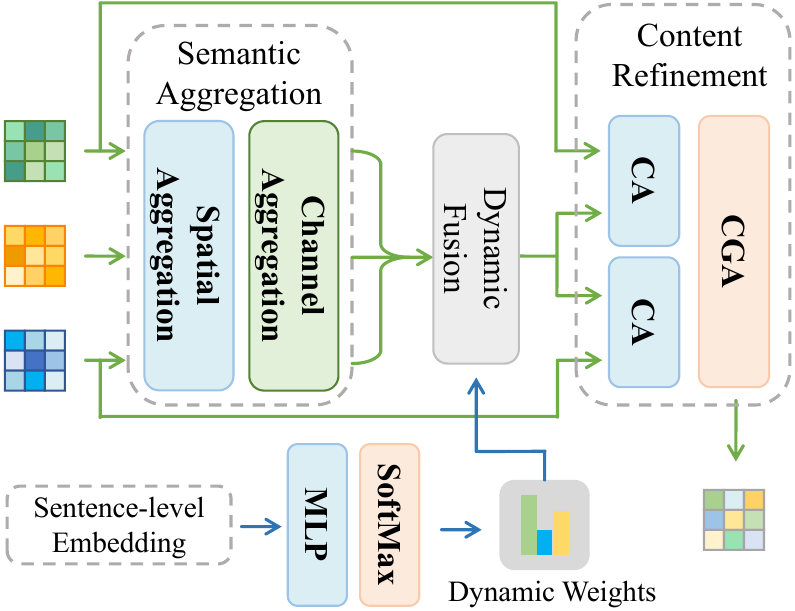} 
\caption{Detailed Architecture of Text-Guided Semantic Fusion Module (TSFM): Progressive Feature Integration via Semantic Aggregation, Text-Guided Dynamic Fusion, and Content Refinement. }
\label{TSFM}
\end{figure}

\subsection{Text-Guided Semantic Fusion Module}

To aggregate the multi-source features of $\mathcal{DF}$, $\mathcal{SF}$, and $\mathcal{MF}$, we dynamically fuse these complementary semantic cues under the guidance of sentence-level embedding, as shown in Fig. \ref{TSFM}.

\textbf{Feature Alignment and Semantic Aggregation.} 
The three input feature maps ($\mathcal{DF}$ from Sec. \ref{sec:method:dpde}, $\mathcal{SF}$ from Sec. \ref{sec:method:sasn},  $\mathcal{MF}$ from Sec. \ref{sec:method:mvin}) first undergo spatial alignment to a unified resolution of $\frac{H}{4} \times \frac{W}{4}$  via bilinear interpolation and residual feature supplementation. 
To harmonize channel dimensions, a multi-layer convolutional projector with GELU activation is employed for each feature, getting aligned features ($\mathcal{SF}^1, \mathcal{DF}^1, \mathcal{MF}^1$).
Consider $\mathcal{DF}^1$ as a representative example, the alignment process can be expressed as:
\begin{equation}\label{W}
\mathcal{DF}^1=P(Down(\mathcal{DF})+Down(P(\mathcal{DF}))),
\end{equation}
where $P(\cdot)$ represents the multi-layer convolutional projector, and $Down(\cdot)$ denotes the downsampling operation.

The aligned features are progressively refined through multiple cascaded aggregation groups, each consisting of a Spatial Aggregation (SA) block followed by a Channel Aggregation (CA) block, as illustrated in the Semantic Aggregation region of Fig. \ref{TSFM}.
This sequential SA-CA architecture can effectively aggregate the intra-feature semantic information, thereby laying a robust foundation for subsequent multi-source feature fusion.
%
Specifically, we aggregate the refined representations ($\mathcal{SF}^2, \mathcal{DF}^2, \mathcal{MF}^2$)  with enriched visual semantic clues\cite{li2022moganet}. 
Taking $\mathcal{DF}^2$ as an illustrative example (processed through one SA-CA stage), the alignment process is formally defined as:
\begin{equation}\label{W}
\mathcal{DF}^2=CA(SA(\mathcal{DF}^1)),
\end{equation}
where $SA(\cdot)$ and $CA(\cdot)$ denote the Spatial Aggregation (SA) block and Channel Aggregation (CA) block, respectively.

\textbf{Sentence Bridge
Routing.}
We employ Sentence-BERT \cite{reimers2019sentence} to extract sentence-level text embedding $\mathcal{S}  = \{S_i\}_{i=1}^K$ that captures holistic scene semantics. 
The fusion weights $\mathcal{W}  = \{W_i\}_{i=1}^K$ are dynamically generated through a multilayer perceptron (MLP) with GELU activation, which modulates the aggregated features to produce text-aware fused representations $\mathcal{TF}  = \{TF_i\}_{i=1}^K$. For each view $i\in\{1,2,\ldots,K\}$, the process is defined as follows:
\begin{equation}\label{W}
W_i=SoftMax(M_2(G(M_1(S_i)))),
\end{equation}
\begin{equation}\label{W}
TF_i=W_i^{SF}\cdot{SF^2_i}+W_i^{DF}\cdot{DF^2_i}+W_i^{MF}\cdot{MF^2_i},
\end{equation}
where $M_1(\cdot)$ and $M_2(\cdot)$ denote MLP layers that first map the text embedding to the hidden space and subsequently project it to the 3-dimensional weight space, $G(\cdot)$ represents the GELU activation function introducing non-linear transformations between the two linear mappings.

This process selectively emphasizes semantically relevant features while suppressing irrelevant components based on linguistic guidance.

\textbf{Content Refinement.} 
The fused features $\mathcal{TF}$ are refined through parallel cross-attention mechanisms with $\mathcal{DF}$ and $\mathcal{SF}$ to generate complementary enhancement features $\mathcal{BF}^{D}$ and $\mathcal{BF}^{S}$, respectively. 
Taking $\mathcal{BF}^{D}$ as an illustrative example, the refinement process is formulated as:
\begin{equation}\label{W}
\mathcal{BF}^{D}=f_c(Cat(WC(\mathcal{TF},\mathcal{DF}),WC(\mathcal{DF},\mathcal{TF}))),
\end{equation}
where $f_c(\cdot)$ denotes a 1×1 convolutional layer, $Cat(\cdot)$ represents concatenation operation along the channel dimension, $WC(\cdot)$ corresponds to the local window cross-attention mechanism.

Subsequently, a Content-Guided Attention module \cite{chen2024dea} further fuses $\mathcal{BF}^{S}$ and $\mathcal{BF}^{D}$ into $\mathcal{BF}^{C}$  according to contextual dependencies.
Finally, the three enhanced features ($\mathcal{BF}^{S}$, $\mathcal{BF}^{D}$,  $\mathcal{BF}^{C}$) are concatenated along the channel dimension and processed by a multi-layer convolutional projector with GELU activation to produce the final refined features $\mathcal{RF}$, which comprehensively integrate multi-source semantic information.

\subsection{Cost Volume-based Decoder}
\hspace*{\parindent} \textbf{Depth Estimation.}
To establish geometric correspondence across views, we construct a cost volume $\mathcal{V}$ through plane-sweep stereo matching using the refined features $\mathcal{RF}$.
The cost volume $\mathcal{V}$ is refined by a lightweight 2D U-Net that takes the channel-wise concatenation of $\mathcal{RF}$ and $\mathcal{V}$ as input. The network predicts a residual feature map, which is added back to the original volume to produce the refined cost representation. 
Depth estimation is achieved by applying softmax normalization across the depth dimension of the refined cost volume, followed by weighted averaging of depth candidates to generate initial per-view depth maps.
A 2D U-Net as a refinement network then takes the concatenation of multi-view images, composite features (obtained by fusing $\mathcal{RF}$ and $\mathcal{CF}$ through a 3×3 convolution), and current depth predictions as input.  This network predicts per-view residual depths that are added to the initial estimates through a residual connection.

\textbf{Gaussian Parameters Prediction.}
The 3D Gaussian centers $\mu$ are derived by unprojecting multi-view depth predictions into world coordinates via camera parameters and merging aligned point clouds.
Opacity $\alpha$ is predicted through two convolutional layers processing softmax-based matching confidence scores from the refined cost volume. 
Covariance $\Sigma$ and color $c$ are predicted by two convolution layers taking concatenated image features, refined cost volume, and original multi-view inputs, with $\Sigma$ parameterized by scaling-rotation decomposition (quaternion-based) and $c$ computed from spherical harmonic coefficients following 3DGS conventions.

\subsection{Loss Function}
Finally, leveraging the predicted 3D Gaussian parameters $\mathcal{G}$, we perform novel view synthesis to generate rendered images $\hat{I}_t$ that are compared with ground-truth images $I_t$.
To optimize the rendering quality, we design a composite loss function $\mathcal{L}_{\text{total}}$ that integrates three key metrics including Mean Squared Error (MSE) loss $\mathcal{L}_{\text{mse}}$, Learned Perceptual Image Patch Similarity (LPIPS) loss $\mathcal{L}_{\text{lpips}}$ \cite{zhang2018unreasonable} and Structure Similarity Index Measure (SSIM) loss $\mathcal{L}_{\text{ssim}}$ \cite{wang2004image}. Formally, the composite loss is defined as:
\begin{equation}
    \mathcal{L}_{\text{total}} = \lambda_{\text{mse}} \mathcal{L}_{\text{mse}} + \lambda_{\text{lpips}} \mathcal{L}_{\text{lpips}} + \lambda_{\text{ssim}} \mathcal{L}_{\text{ssim}},
\end{equation}
where $\mathcal{L}_{\text{mse}}$ measures pixel-wise differences between rendered and ground-truth images, $\mathcal{L}_{\text{lpips}}$ captures perceptual feature discrepancies, and $\mathcal{L}_{\text{ssim}}$ evaluates local structural consistency. The hyperparameters $\lambda_{\text{mse}}=1.0$, $\lambda_{\text{lpips}}=0.05$, and $\lambda_{\text{ssim}}=0.03$ balance the contributions of each loss term.

\begin{table}[t]
  \centering
  \caption{Training hyperparameters for the proposed TextSplat model across different training stages and datasets.}
  \renewcommand{\arraystretch}{1.0}
    \begin{tabular}{c|ccc}
    \toprule
    & Batch Size & Learning Rate & Iteration \\
    \midrule
    Train Stage1 & 4     & 5e-5  & 50,000 \\
    Train Stage2 & 3     & 2e-5  & 100,000 \\
    Train Stage3 (ACID) & 2     & 5e-6  & 50,000 \\
    Train Stage3 (RE10K) & 2     & 5e-6  & 150,000 \\
    \bottomrule
    \end{tabular}%
  \label{hyperparameters}%
\end{table}

\section{EXPERIMENTS}

\subsection{Experiment Setup}

\begin{table*}[htp]
  \centering
  \setlength{\tabcolsep}{10pt}
  \caption{Quantitative Comparison with State-of-the-Art Methods on RealEstate10K and ACID Datasets. Bold indicates best performance, underline denotes second best. Our TextSplat demonstrates achieves competitive results against 4 NeRF-based and 5 3DGS-based approaches. Bold indicates the best performance and underlined indicates second place.}
  \renewcommand{\arraystretch}{1.0}
  \begin{tabular}{c|c|c|ccc|ccc}
    \toprule
    \multicolumn{3}{c|}{\multirow{2}[4]{*}{Method}} & \multicolumn{3}{c|}{RealEstate10K \cite{zhou2018stereo}} & \multicolumn{3}{c}{ACID \cite{liu2021infinite}} \\
\cmidrule{4-9}    \multicolumn{3}{c|}{} & PSNR↑ & SSIM↑ & LPIPS↓ & PSNR↑ & SSIM↑ & LPIPS↓ \\
    \midrule
    \multirow{4}[2]{*}{NeRF-based} & PixelNeRF \cite{yu2021pixelnerf} & CVPR 2021 & 20.43  & 0.589  & 0.550  & 20.97 & 0.547 & 0.533  \\
          & GPNR \cite{suhail2022generalizable}  & ECCV 2022 & 24.11  & 0.793  & 0.255  & 25.28 & 0.764 & 0.332  \\
          & AttnRend \cite{du2023learning} &  CVPR 2023 & 24.78  & 0.820  & 0.213  & 26.88 & 0.799 & 0.218  \\
          & MuRF \cite{xu2024murf}  & CVPR 2024 & 26.10  & 0.858  & 0.143  & 28.09 & 0.841 & 0.155  \\
    \midrule
    \multirow{6}[4]{*}{3DGS-based} & PixelSplat \cite{charatan2024pixelsplat} & CVPR 2024 & 25.89  & 0.858  & 0.142  & 28.14 & 0.839 & 0.150  \\
          & MVSplat \cite{chen2024mvsplat} & ECCV 2024 & 26.39  & 0.869  & 0.128  & \underline{28.25} & 0.843 & 0.144  \\
          & MVSplat360 \cite{chen2024mvsplat360} & NeurIPS 2024 & 26.41  & 0.869  & 0.126  & /     & /     & / \\
          & TranSplat \cite{zhang2024transplat} & AAAI 2025 & \underline{26.69}  & \underline{0.875}  & \underline{0.125}  & \textbf{28.35} & \underline{0.845} & \textbf{0.143} \\
          & Omni-Scene \cite{wei2024omni} & CVPR 2025 & 26.19  & 0.865  & 0.131  & /     & /     & / \\
\cmidrule{2-9}          & TextSplat & Ours  & \textbf{27.08} & \textbf{0.885} & \textbf{0.121} & \textbf{28.35} & \textbf{0.849} & \underline{0.144} \\
    \bottomrule
    \end{tabular}%
  \label{Main Compare}%
\end{table*}%

\begin{table}[ht]
  \centering
  \setlength{\tabcolsep}{8pt}
  \caption{Quantitative Comparison on Cross-dataset Generalization. The consistently superior performance across all metrics demonstrates the strong generalization capability of our method.}
  \renewcommand{\arraystretch}{1.0}
    \begin{tabular}{c|c|ccc}
    \toprule
    \multicolumn{2}{c|}{\multirow{2}[4]{*}{Method}} & \multicolumn{3}{c}{RealEstate10K→ACID} \\
\cmidrule{3-5}    \multicolumn{2}{c|}{} & PSNR↑ & SSIM↑ & LPIPS↓ \\
    \midrule
    PixelSplat & CVPR 2024 & 27.64  & 0.830  & 0.160  \\
    MVSplat & ECCV 2024 & 28.15  & 0.841  & \underline{0.147}  \\
    TranSplat & AAAI 2025 & \underline{28.17}  & \underline{0.842}  & \textbf{0.146} \\
    \midrule
    TextSplat & Ours  & \textbf{28.27} & \textbf{0.847} & \textbf{0.146} \\
    \bottomrule
    \end{tabular}%
  \label{Cross-dataset}%
\end{table}%

\hspace*{\parindent} \textbf{Datasets.}
We evaluate the reconstruction performance on two large-scale benchmarks: RealEstate10K \cite{zhou2018stereo} and ACID \cite{liu2021infinite}. 
Specifically, RealEstate10K contains 74,766 YouTube real estate videos (67,477 training, 7,289 testing), while ACID comprises 13,047 drone-captured nature scenes (11,075 training, 1,972 testing).  
This combination covers diverse environments from structured indoor/outdoor spaces to unstructured natural landscapes.
Both datasets provide camera intrinsic/extrinsic parameters for each frame, calibrated using the Structure-from-Motion (SfM) algorithm \cite{schonberger2016structure}.

\textbf{Evaluation Criteria.}
We evaluate the generation quality of RGB images using three complementary metrics: pixel-level PSNR for reconstruction quality, patch-level SSIM \cite{wang2004image} for structural consistency, and feature-level LPIPS \cite{zhang2018unreasonable} for perceptual quality, providing a multi-scale assessment of fidelity, coherence, and visual realism.

\textbf{Implementation Details.}
To ensure fair comparisons across studies, we aligned our training protocols with existing works. Specifically, all experiments were performed using two input perspective images with 256×256 resolution. The model was implemented in PyTorch \cite{paszke2019pytorch} and leveraged a CUDA-accelerated 3DGS renderer to optimize computational efficiency.
All models were trained on 6×A6000 Ada GPUs for 200,000 iterations on ACID and 300,000 iterations on RealEstate10K using the Adam \cite{kingma2014adam} optimizer with cosine annealing learning rate scheduling.

\textbf{Model Initialization.}
The Multi-View Interaction Network and Cost Volume-based Decoder are initialized with pre-trained parameters from MVSplat \cite{chen2024mvsplat}; the Diffusion Prior Depth Estimator inherits parameters from TADP \cite{kondapaneni2024text}; while the Semantic Aware Segmentation Network adopts weights from OMG-SEG \cite{li2024omg}.

\textbf{Training Stages.}
We adopt a three-stage progressive training strategy to balance optimization stability and computational efficiency, with key hyperparameters summarized in Table~\ref{hyperparameters}:
1) Exclusive fine-tuning of the Text-Guided Semantic Fusion Module.
2) Joint optimization of the fusion module, Multi-View Interaction Network, and Cost Volume-based Decoder.
3) Full model fine-tuning with frozen encoders in Diffusion Prior Depth Estimator and Semantic Aware Segmentation Network.
Notably, we employ an evolving fusion mechanism: direct feature summation in Stages 1-2 transitions to dynamic fusion with text embedding in Stage 3.

\textbf{Description Generation.}
To reduce computational overhead, video frames in the training set are chronologically divided into five sub-sequences.  The visual description for the first frame of each sub is shared across all frames within the same group. 
During testing, independent visual descriptions are generated for each input view image.
All visual descriptions were produced using the pre-trained Qwen2-VL model \cite{wang2024qwen2}.

\begin{figure*}[t]
\centering
\includegraphics[width=0.88\linewidth]{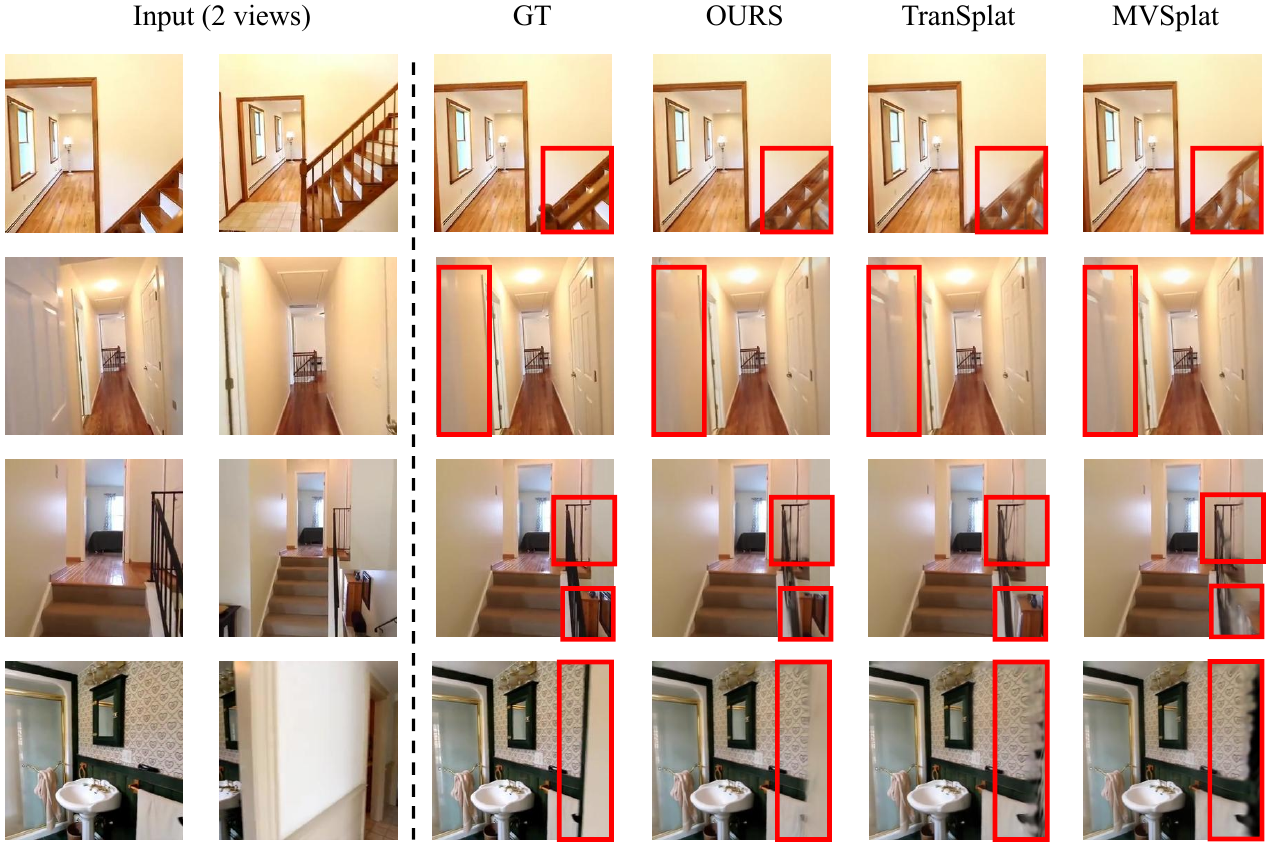} 
\caption{Qualitative comparison of different models. TextSplat achieves superior rendering quality in challenging regions, benefiting from the high-quality feature representations generated by our proposed Text-Guided Semantic Fusion Module.}
\label{qualitative comparisons}
\end{figure*}

\subsection{Comparison With State-of-The-Art}

To comprehensively evaluate the proposed TextSplat framework, we conduct both quantitative and qualitative comparisons with state-of-the-art models. Furthermore, we perform cross-dataset evaluations to validate its generalization capability.


\textbf{1) Quantitative evaluation:}
As shown in Table \ref{Cross-dataset}, We perform quantitative evaluations against 9 state-of-the-art models, including 4 NeRF-based methods (PixelNeRF \cite{yu2021pixelnerf}, GPNR \cite{suhail2022generalizable}, AttnRend \cite{du2023learning} and MuRF \cite{xu2024murf}) and 5 3DGS-based approaches (PixelSplat \cite{charatan2024pixelsplat}, MVSplat \cite{chen2024mvsplat}, MVSplat360 \cite{chen2024mvsplat360}, TranSplat \cite{zhang2024transplat}, and Omni-Scene \cite{wei2024omni}).

\textit{Performance on RealEstate10K.} 
Our TextSplat achieves 27.08 dB PSNR, 0.885 SSIM, and 0.121 LPIPS on RealEstate10K benchmark, demonstrating best performance when compared with competitors. 
Compared to MuRF \cite{xu2024murf} - the best-performing NeRF-based method, TextSplat outperforms by a significant margin of +0.98 dB in PSNR and +0.027 in SSIM. 
Furthermore, when contrasted with TranSplat \cite{zhang2024transplat} - the leading GGS approach, our method demonstrates superior performance with improvements of +0.39 dB in PSNR and +0.01 in SSIM. 

\textit{Performance on ACID.}
For the ACID dataset, TextSplat achieves 28.35 dB PSNR and 0.849 SSIM, ranking first among all competing methods. 
While TranSplat \cite{zhang2024transplat} reports a marginally better LPIPS score (0.143 vs. 0.144), our approach achieves a +0.004 improvement in SSIM, demonstrating enhanced structural fidelity in reconstructing complex natural scenes with rich geometric details. 
This enhancement highlights the effectiveness of our Text-Guided Semantic Fusion Module in maintaining perceptual consistency during scene reconstruction.

\begin{figure*}[h]
\centering
\includegraphics[width=0.92\linewidth]{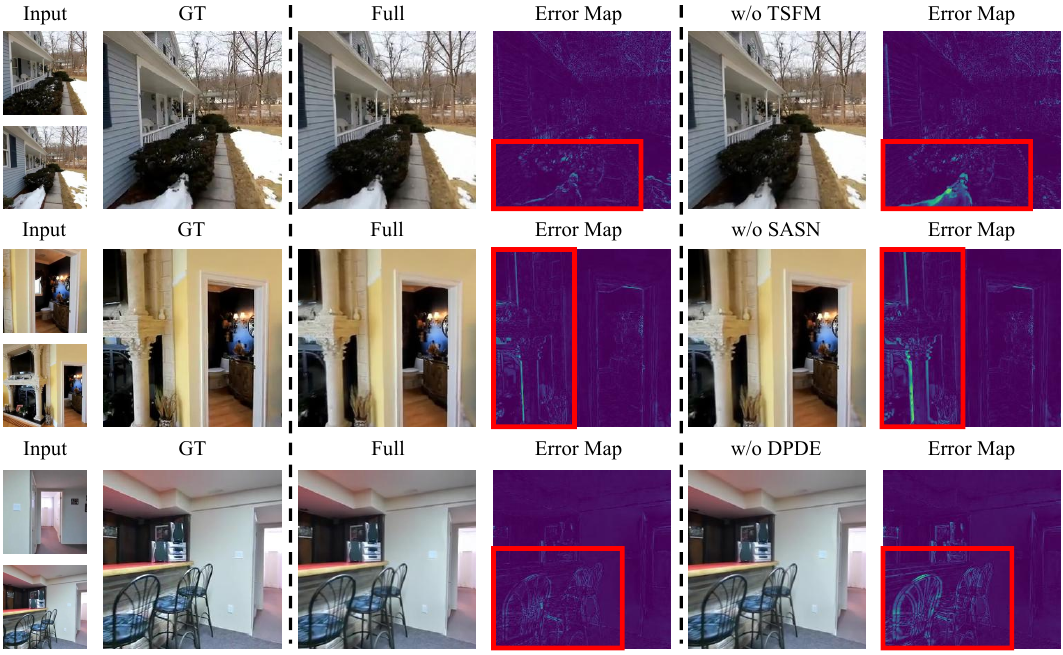} 
\caption{Qualitative comparison of ablation models. Removing the TSFM module (w/o TSFM) impairs prediction of irregular-shaped object geometries; eliminating the SASN module (w/o SASN) introduces part-level semantic ambiguity; discarding the DPDE module (w/o DPDE) degrades spatial localization in complex occlusion scenarios. Colored error maps between rendered outputs and ground truth are visualized to highlight performance differences.}
\label{ab comparisons}
\end{figure*}

\begin{table}[t]
  \centering
  \setlength{\tabcolsep}{12pt}
  \caption{Ablation Study of Core Components for TextSplat on RealEstate10K.}
  \renewcommand{\arraystretch}{1}
    \begin{tabular}{c|ccc}
    \toprule
    Setup & PSNR↑ & SSIM↑ & LPIPS↓ \\
    \midrule
    w/o TSFM & 26.82  & 0.880  & 0.124  \\
    w/o SASN & 26.83  & 0.881  & 0.124  \\
    w/o DPDE & 26.58  & 0.876  & 0.127  \\
    \midrule
    Full  & \textbf{27.08} & \textbf{0.885} & \textbf{0.121} \\
    \bottomrule
    \end{tabular}%
  \label{Components Ablation}%
\end{table}%

\textbf{2) Qualitative Evaluation:}
Fig. \ref{qualitative comparisons} presents the visual comparisons across four challenging scenarios.
The first row shows a bright stairwell scene with significant occlusions between stair steps and railings. Due to the large distance between input views and the complex spatial relationships caused by occlusions, accurately reconstructing this scene is highly challenging. While TranSplat and MVSplat produce extensive blurring and distortions around the railings, our TextSplat achieves significantly better reconstruction by preserving the fine details and spatial relationships of the stairs and railings.
The second row highlights a low-contrast hallway where walls and doors confuse other methods into generating ghosting artifacts, whereas our approach preserves wall integrity.
The third row demonstrates a multi-floor complex space. TranSplat incorrectly assigns the white color of railing bases to adjacent floor cabinets, while MVSplat introduces severe blur in the lower-floor area. In contrast, our method accurately reconstructs both the color boundaries and the intricate railing shapes.
The final row presents a large field-of-view occlusion scenario. Existing methods mispredict wall positions, resulting in black voids, whereas TextSplat accurately reconstructs the walls with precise spatial localization.
These results validate our framework's superiority in resolving geometry-semantic ambiguities under sparse-view constraints.

\textbf{3) Cross-dataset Generalization:}
In Table \ref{Cross-dataset}, we evaluate the generalization capability of the proposed TextSplat through zero-shot cross-dataset validation. The model is initially trained on the RealEstate10K dataset and subsequently tested on the ACID benchmark in a zero-shot setting, demonstrating robust performance without domain-specific fine-tuning.
The results demonstrate TextSplat's superior capability in generalizing to unseen domains, achieving 28.27 dB PSNR and 0.847 SSIM that surpass all competing methods, while maintaining an LPIPS score of 0.146 comparable to the state-of-the-art TranSplat.

\subsection{Ablation Analysis}
We conduct comprehensive ablation studies on the RealEstate10K dataset, covering both core components and progressive training strategies.

\begin{table}[t]
  \centering
  \setlength{\tabcolsep}{10pt}
  \caption{Ablation Study of Training Stages on RealEstate10K, where TS denotes Training Stage.}
  \renewcommand{\arraystretch}{1.0}
    \begin{tabular}{ccc|ccc}
    \toprule
    TS1   & TS2   & TS3   & PSNR↑ & SSIM↑ & LPIPS↓ \\
    \midrule
    $\checkmark$    & $\times$     & $\times$      & 26.55 & 0.874 & 0.128 \\
    $\checkmark$     & $\checkmark$     &  $\times$     & 26.75 & 0.879 & 0.124 \\
    $\times$      &  $\times$     & $\checkmark$     & 26.86 & 0.881 & 0.123 \\
    $\checkmark$     & $\checkmark$     & $\checkmark$     & \textbf{27.08} & \textbf{0.885} & \textbf{0.121} \\
    \bottomrule
    \end{tabular}%
  \label{Stages Ablation}%
\end{table}%

\textbf{1) Effect of Core Components:} Table \ref{Components Ablation} demonstrates the impact of core components on text-driven 3D reconstruction performance. 
This quantitative analysis is further corroborated by qualitative comparisons in Figure \ref{ab comparisons}.
The TSFM module contributes 0.26 dB PSNR gain (27.08→26.82 when removed), validating the effectiveness of text-guided dynamic fusion. 
Specifically, the error map in Figure \ref{ab comparisons} shows that removing TSFM leads to inaccurate predictions of irregular-shaped objects like snow piles in outdoor scenes, while the full model reconstructs their complex geometries more precisely. 
Removing SASN causes SSIM drop (0.885→0.881), indicating explicit semantic supervision complements geometric details.
This is visually evident in indoor scenes where the absence of SASN results in ambiguous boundaries between columns and background cabinets, as highlighted by the error map's red regions. 
DPDE shows the most significant impact on geometry-semantic consistency, with LPIPS increasing by 0.006 (0.121→0.127) when removed.
The error visualization confirms its critical role in spatial localization under occlusion—removing DPDE causes misalignment in occluded areas like furniture arrangements, whereas the full model accurately determines their spatial positions through diffusion priors.

\textbf{2)  Effect of Training Strategies:} Table \ref{Stages Ablation} demonstrates the effectiveness of our progressive training strategy in optimizing multimodal feature fusion. The results reveal that each training stage contributes distinct improvements. 
Stage 1 (TS1) establishes foundational semantic alignment, achieving PSNR 26.55 and SSIM 0.874. 
Adding Stage 2 (TS2) joint optimization with the Multi-View Interaction Network and Cost Volume Decoder, improving geometric consistency and boosting PSNR to 26.75 (+0.20 dB) and SSIM to 0.879 (+0.05).  
The full pipeline incorporating Stage 3 (TS3) enables full model fine-tuning, achieving the best performance (PSNR 27.08, SSIM 0.885, LPIPS 0.121).
Isolated Stage 3 training yields suboptimal results (26.86 PSNR, -0.22↓), underscoring the necessity of progressive optimization.

\section{Conclusion}
In conclusion, this paper introduces TextSplat, the first text-driven Generalizable Gaussian Splatting framework that enhances geometry–semantic consistency in sparse-view scenarios through dynamic multi‑source semantic fusion. 
By integrating a Diffusion Prior Depth Estimator, Semantic Aware Segmentation Network, and Multi-View Interaction Network with the Text-Guided Semantic Fusion Module, TextSplat achieves state-of-the-art reconstruction quality on RealEstate10K and ACID datasets.
Future work will investigate more efficient text-driven dynamic fusion mechanisms to enhance compatibility in complex scenarios, extending their application to diverse 3D visual tasks.



\bibliographystyle{ACM-Reference-Format}
\balance
\bibliography{sample-base}

\appendix

\section{Qualitative Comparison of 3D Reconstruction Performance}

\begin{figure}[h]
\centering
\includegraphics[width=0.95\linewidth]{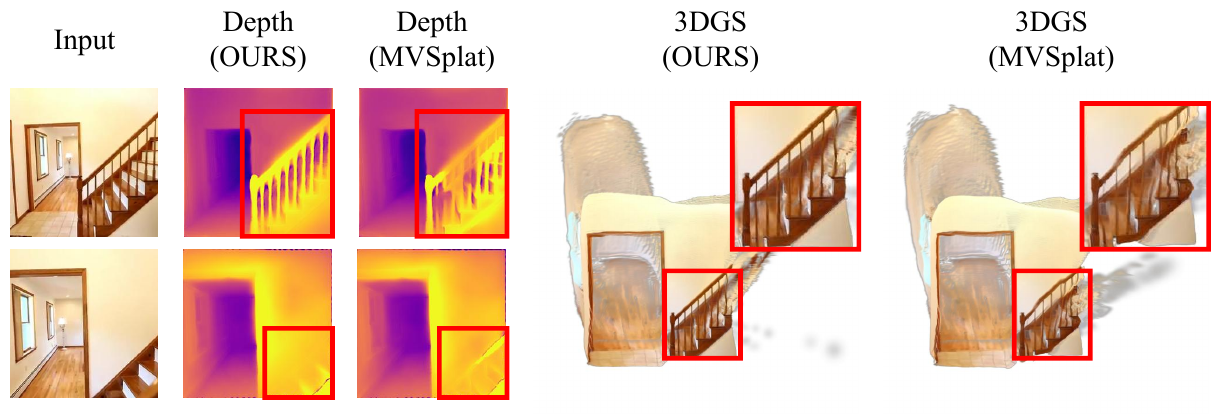} 
\caption{Qualitative Comparison of Depth Estimation 3D Reconstruction Performance. }
\label{3d_comparisons}
\end{figure}

As shown in Fig. \ref{3d_comparisons}, we have conducted qualitative visualization for depth estimation and 3D reconstruction.
Specifically, within the stairwell region, our TextSplat approach achieves significantly more accurate depth estimation than the MVSplat. This superior performance translates to enhanced preservation of fine structural details within the reconstructed 3D Gaussian representations.

\section{Feature Visualization}

\begin{figure}[b]
\centering
\includegraphics[width=0.95\linewidth]{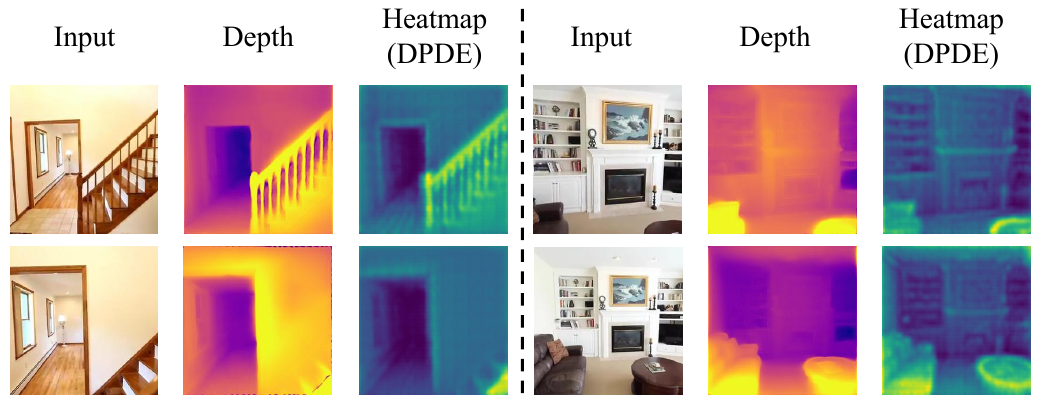} 
\caption{Feature Visualizations of the DPDE. }
\label{depth_feature}
\end{figure}

As shown in Fig. \ref{depth_feature}, we have conducted feature visualizations of the DPDE outputs, confirming that diffusion-based depth representations capture rich geometric patterns. These visualizations demonstrate that our method consistently generates high-fidelity depth cues across diverse scenes.

\section{Limitation}

\begin{figure}[b]
\centering
\includegraphics[width=0.95\linewidth]{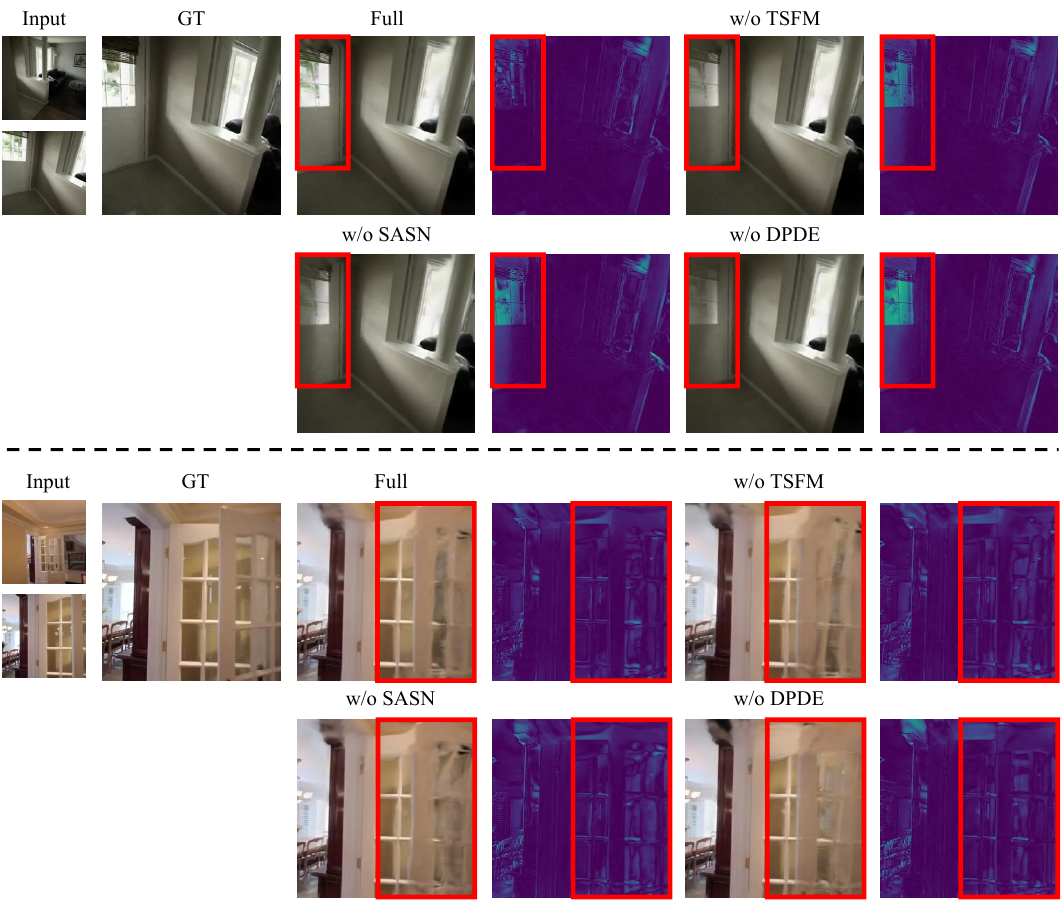} 
\caption{Failure Cases on Transparent Materials. }
\label{false}
\end{figure}

Despite the significant improvements achieved by our TextSplat framework in geometry-semantic consistency and reconstruction quality, certain limitations remain that warrant further investigation.
As shown in Fig. \ref{false}, while our model effectively enhances reconstruction performance through its Text-Guided Semantic Fusion Module, challenges persist in accurately reconstructing scenes containing transparent materials such as glass doors and windows, as well as objects behind them.
This limitation highlights the need for more sophisticated modeling of transparent materials and their interactions with surrounding objects. Future work will explore advanced material-aware priors and specialized loss functions to address these issues and further enhance scene fidelity.

\end{document}